# Twins Recognition Using Hierarchical Score Level Fusion

Cihan AKIN, Umit KACAR, and Murvet KIRCI

Department of Electrical and Electronic Engineer, Istanbul Technical University, Istanbul, Turkey

**Abstract**

With the development of technology, the usage areas and importance of biometric systems have started to increase. Since the characteristics of each person are different from each other, a single model biometric system can yield successful results. However, because the characteristics of twin people are very close to each other, multiple biometric systems including multiple characteristics of individuals will be more appropriate and will increase the recognition rate. In this study, a multiple biometric recognition system consisting of a combination of multiple algorithms and multiple models was developed to distinguish people from other people and their twins. Ear and voice biometric data were used for the multimodal model and 38 pair of twin ear images and sound recordings were used in the data set. Sound and ear recognition rates were obtained using classical (hand-crafted) and deep learning algorithms. The results obtained were combined with the hierarchical score level fusion method to achieve a success rate of 94.74% in rank-1 and 100% in rank -2.

## 1. Introduction

With the increasing of twins' population biometric systems should work successfully with twins' datasets. So, there are studies about twins and multi modal biometric systems.

Distinguishing of twins has been studied by using the facial marks of people. Mole, freckle, freckle group, lightened patch, darkened patch, birthmark, splotches, raised skin, scar, pockmark and pimple are used as facial marks. 178 people consisting of 89 pairs of twins and 477 images were used as dataset [1].

Distinguishing of twins from each other has been studied by using face recognition algorithms [2, 3].

Lakshmi Priya and Pushpa Rani designed a multi modal biometric system by using fingerprint, lip print and face images. 214 pairs of twins were used as dataset [4].

Bayan Omar Mohammed and Siti Mariyam Shamsuddin designed a multi modal biometric recognition system by using fingerprint and hand writing. There are 20 pairs of twins in dataset [5, 6].

Cross modal learning and shared representation learning algorithms were studied in a study on multi modal deep learning. CUAVE and AV Letters datasets were used for this study [7].

## 2. Multi Algorithm Multi Modal

Audio-Visual Twins Database (AVTD) was used in this study. This dataset consists of voice records and left-right ear images of 39 pairs of twins but, there isn't right ear image of one person in dataset. So, voice records and face images of 38 pairs of twins were used in this study. There are 3 voice records for one person. 2 of voice records were used as training set and 1 of voice records were used as test set. There are 2 ear images for one person. Left ear image was used as training set and right ear image was used for test set.

## 3. Voice Recognition Algorithms

### 3.1. Mel Frequency Cepstral Coefficients

Mel Frequency Cepstral Coefficients (MFCC) is used commonly for feature extraction in voice recognition systems. It models human's hear perception. First, pre-emphasis filter is applied to voice signal. Filtered signal is divided into small frames and "hamming window" is applied to each frame. Fast Fourier Transform is applied for conversion from time domain to frequency domain. Then, "mel filter bank" is applied to FFT results and finally MFCC is obtained by using Discrete Cosine Transform.

### 3.2. Long Short-Term Memory Network (LSTM)

Long Short-Term Memory (LSTM) was developed to solve the long-term dependency of Recurrent Neural Network. LSTM was proposed in 1997 by Hochreiter and Schmidhuber. LSTM structure is shown in "Fig. 1".

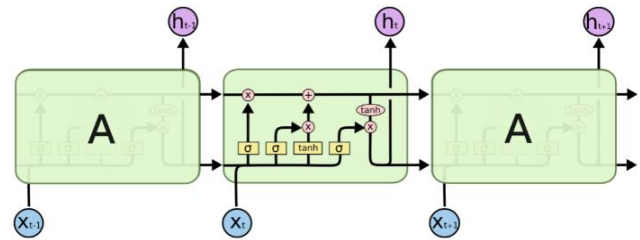

**Fig. 1.** Long Short-Term Memory

LSTM has three layers as forget gate layer, input gate layer and output gate layer.

Forget gate decides how much of the previous data will be forgotten and how much of the previous data will be used in next steps. The result of this gate is in the range of 0-1 while "0" forgets the previous data, "1" uses the previous data. Forget gate layer can be modelled as in equation 1.

$$f_t = \sigma * (W_f * [h_{t-1}, x_t] + b_f) \quad (1)$$

Second layer is input gate layer which consist of input gate layer and tanh layer. New data is is obtained in this layer. Unnecessary part of input data is filtered with a sigmoid function and then new possible data is determined with a tanh function. Multiplication of sigmoid function's result and tanh layer's result is added to cell state to update cell state and new cell state is obtained. Input gate layer can be modelled as in equation 2, equation 3 and equation 4.

$$i_t = \sigma * (W_i * [h_{t-1}, x_{t-1}] + b_i) \quad (2)$$

$$\tilde{C}_t = \tanh * (W_c * [h_{t-1}, x_{t-1}] + b_c) \quad (3)$$

$$C_t = f_t * C_{t-1} + i_t * \tilde{C}_t \quad (4)$$

In output gate layer, cell state is filtered by using tanh function and input data is filtered with a sigmoid function. Multiplication of sigmoid function's result and tanh layer's result becomes output data. Output gate layer can be modeled as in equation 6 and equation 7.

$$o_t = \sigma \times (W_o \times [h_{t-1}, x_{t-1}] + b_o) \quad (6)$$

$$h_t = o_t \times \tanh(C_t) \quad (7)$$

### 3.3. Dynamic Time Warping (DTW) Algorithm

People say the same words fast or slowly, in different times. So, voice signal is a time dependent signal. Dynamic Time Warping algorithm (DTW) was developed (Doddington 1971) to solve time dependency problems [8]. DTW is a good algorithm for voice recognition systems that solves time dependency problem.

This algorithm aims to create a warping function according to the minimum distance between the training set and test set. In this way, the best matching between the test set and the training set is determined by solving time dependency problem of voice signal. DTW can be modelled as in equation 8 and equation 9.

$$d = d(E_i, T_j) = \sqrt{(E_i - T_j)^2} \quad (8)$$

$$D(i,j) = min[D(i-1, j-1), D(i-1, j), D(i, j-1)] + d \quad (9)$$

## 4. Ear Recognition Algorithms

### 4.1. DenseNet

DenseNet, is a method of Convolutional Neural Networks (CNN). An image is processed many convolutional operations in CNN and high-level features are obtained. Output of one layer becomes the input of another layer in CNN structure.

In DenseNet network structure, one layer gets the feature vectors from output of all previous layers and transfers these feature vectors to next layers. In this way, network structure is smaller because each layer has the all previous feature vectors. This is an advantage of DenseNet as memory and speed according to CNN structure [11].

### 4.2. Histogram of Oriented Gradients (HOG)

Histogram of Gradients (HOG) was proposed for human detection by Dalas and Triggs. Gradient of image is calculated in the first step. Then colour and density values of image are filtered for calculation process. In second step image is divided into small cells and each cell's histograms of quantized gradient orientations are calculated. The pixels which have higher gradient magnitudes, affects histogram bins more than the pixels which have lower gradient magnitudes. Then neighbour cells are grouped and normalized by accounting potential changes in contrast and illumination. Normalization process is applied by sliding over the entire image and there can be overlap in some groups. Finally, HOG that can be used for recognition, is obtained [12].

## 5. Hierarchical Score Level Fusion

Fusion aims to achieve the best results by combining the results obtained by different algorithms. In the literature, there are 5 types of fusion level methods, including sensor, attribute, score, sequence and decision level used in the fusion process [9].

The score level fusion method includes 4 different score level fusion methods as classical, hierarchical, cascade and hybrid. Hierarchical score level fusion mechanism is used in this study and is shown in "Fig. 2". The hierarchical score level fusion mechanism is ideal for combining multiple algorithms and multiple models. Fusion 1 and fusion 2 represent multiple algorithms and fusion 3 represents multiple models.

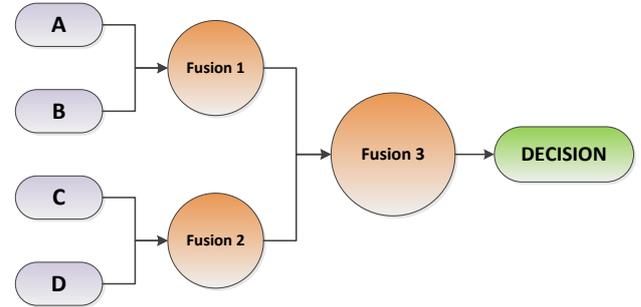

**Fig. 2.** Hierarchical score level fusion mechanism

## 6. Experimental Results

The block diagram of multi modal multi algorithm biometric system is shown in "Fig. 3".

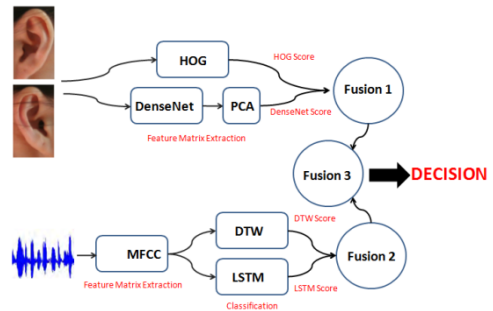

**Fig. 3.** Biometric system block diagram

In ear recognition system, feature matrices were extracted by using the DenseNet network. However, due to the large size of the feature matrix, size reduction was performed by using Principal Component Analysis (PCA). The recognition rate obtained from the first modality was 51.32%. HOG was used as the second modality and the recognition rate of this modality was calculated as 38.16%. Score matching matrices were obtained by using "Manhattan" metric space for both modalities. Then two modalities combined by using score levels were fusion within the scope of the multi-algorithm. As a result of fusion, 52.63% recognition rate was found for ear recognition.

In voice recognition system, feature matrices were extracted by using MFCC algorithm. In the first modality, classification was made by using DTW and the recognition rate was 90.79%. The second modality was LSTM and the recognition rate was calculated as 61.84%.

After obtaining recognition score matrices for ear and voice, "tanh" normalization was used as the normalization process. Because all score must be in the same scale. This normalization process was also performed in the multi algorithm applications for ear and voice. In summary, hierarchical score level fusion was used for firstly for multi algorithm and secondly for multi modal.

The result of score level fusion method for two different ear recognition algorithms was obtained as %52,63 in rank-1 and %67,11 in rank-2.

The result of score level fusion method for two different voice recognition algorithms was obtained as %93,42 in rank-1 and %97,37 in rank-2.

Cumulative matching characteristic (CMC) curves of ear recognition algorithm and voice recognition algorithm are shown in "Fig. 4".

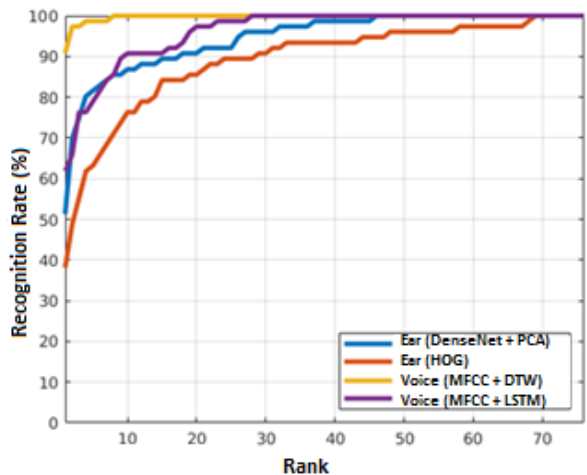

**Fig. 4.** Ear and voice recognition rates

Two score matrices were obtained that one score matrix is the result of voice recognition algorithms and one score matrix is the result of ear recognition algorithms. Then, two score matrices were applied score level fusion method. As a result, recognition rate was %94,74 in rank-1 and %100 in rank-2.

In summary, the following operations are listed:
- Ear biometric system (multi-algorithm)
- Voice biometric system (multi-algorithm)
- Ear and voice recognition system (multi-modal)
- Hierarchical score level fusion mechanism

The cumulative matching characteristic curves of the fusion results are presented in "Fig. 5" and all recognition results are presented in Table 1.

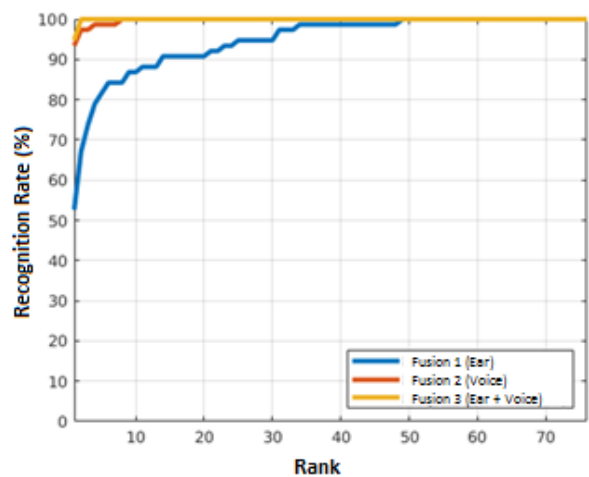

**Table 1.** Classification results

| Biometric Systems | Fusion Weight | Rank-1 | Rank -2 | Rank -5 | AUC[1] |
|---|---|---|---|---|---|
| Ear (Gabor + DCVA) [10] | 0.15 | 43,4 | 55,3 | 71,1 | - |
| Voice (MFCC + DTW) [10] | 0.85 | 80,3 | 89,5 | 96,1 | - |
| Fusion-1 (Ear + Voice) [10] | - | 81,6 | 94,7 | **100** | - |
| Ear (HOG) | 0.21 | 38,16 | 48,68 | 63,16 | 89,22 |
| Ear (DenseNet +PCA) | 0.79 | 51.32 | 69,74 | 81,58 | 94,80 |
| Voice (MFCC + DTW) | 0.98 | 90,79 | 97,37 | 98,68 | 99,80 |
| Voice (MFCC + LSTM) | 0.02 | 61,84 | 65,79 | 78,95 | 96,31 |
| Fusion -1 (Ear) | 0.14 | 52,63 | 67.11 | 81,58 | 94,87 |
| Fusion -2 (Voice) | 0.86 | 93,42 | 97,37 | 98,68 | 99,82 |
| Fusion -3 (Ear + Voice) | - | **94,74** | **100** | **100** | **99,97** |

## 7. Conclusion

In the previous study [10] multi-model voice and ear recognition system had designed. Ear recognition system had only one algorithm and voice recognition system had only one algorithm. Recognition rate had been 81.6% for rank-1 and 94.7% for rank-2. 100% recognition rate was obtained in rank-5.

In this study, both ear recognition system and voice recognition system had two algorithms and firstly these two algorithms are applied fusion process for ear and voice modals. Then, the results of ear and voice recognition system were applied fusion process. As a result, the recognition rate was improved by 10% compared to the previous study. In this study, multi algorithm multi modal biometric system was realized by using hierarchical score level fusion mechanism and recognition rate is 94.74% in rank-1 and 100% in rank-2. Since our data set is small and controlled, the recognition rate is high, and in the following studies it will be studied with a large and uncontrolled dataset.

---

[1]Area under curve